\documentclass{article}
\usepackage{spconf,amsmath,graphicx}
\usepackage{multirow}
\usepackage{amsfonts}

\title{Visual Localization Using Sparse Semantic 3D Map}
\name{Tianxin Shi$^{\star \dagger}$ \qquad Shuhan Shen$^{\star \dagger}$ \qquad Xiang Gao$^{\star \dagger}$ \qquad Lingjie Zhu$^{\star \dagger}$\thanks{This work was supported by the Natural Science Foundation of China under Grants 61632003, 61873265.}}
\address{$^{\star}$NLPR, Institute of Automation, Chinese Academy of Sciences, Beijing 100190, China\\
$^{\dagger}$University of Chinese Academy of Sciences, Beijing 100049, China}

\begin{document}
%
\maketitle
\begin{abstract}
Accurate and robust visual localization under a wide range of viewing condition variations including season and illumination changes, as well as weather and day-night variations, is the key component for many computer vision and robotics applications. Under these conditions, most traditional methods would fail to locate the camera. In this paper we present a visual localization algorithm that combines structure-based method and image-based method with semantic information. Given semantic information about the query and database images, the retrieved images are scored according to the semantic consistency of the 3D model and the query image. Then the semantic matching score is used as weight for RANSAC's sampling and the pose is solved by a standard PnP solver. Experiments on the challenging long-term visual localization benchmark dataset demonstrate that our method has significant improvement compared with the state-of-the-arts.
\end{abstract}

\begin{figure*}[htb]
	\centering  
	\includegraphics[width=16cm,height=4.2cm]{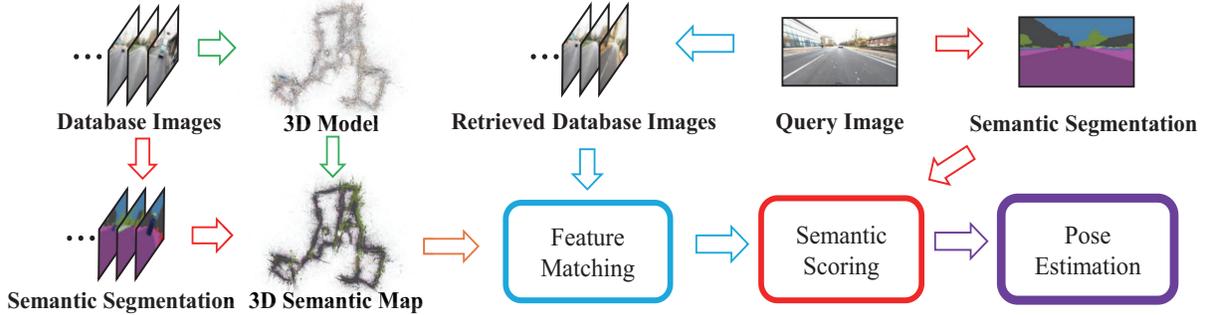}  
	\caption{Flowchart of the localization pipeline proposed in the paper. } 
	\label{fig:1}   
\end{figure*}

\section{Introduction}
\label{sec:intro}

Visual localization plays a central role in many computer vision and robotics applications, such as loop closure detection \cite{7989618} and re-localization \cite{li2010location} in SLAM \cite{mur2015orb}, Structure-from-Motion (SfM) \cite{schonberger2016structure}, augmented reality \cite{castle2008video}, and autonomous vehicles, which has recently drawn a lot of attentions.

Currently, there are three main types of approaches to the localization problem. \textbf{3D structure-based methods} \cite{sattler2017efficient, liu2017efficient, zeisl2015camera, svarm2017city} establish matches between 2D local features in the query image and 3D points in the model and use these correspondences to recover the 6DoF camera pose. \textbf{2D image-based localization methods} \cite{arandjelovic2016netvlad, lowry2016visual, sattler2016large, torii201524} cast the localization problem as an image retrieval problem. The camera pose of the most relevant retrieved database image is used as an approximation to the query image position. \textbf{Learning-based methods} \cite{chen2017deep, cao2013graph, kendall2015posenet} use the learned model for descriptor learning \cite{chen2017deep} or use the model to regress the camera pose directly\cite{kendall2015posenet}. All of the three types of approaches have advantages and disadvantages. The structure-based approach is more accurate than the other two methods, but it becomes very time-consuming as the scale of the scene increases. Although image-based methods are faster, they are often considered inaccurate since they only approximate the positions of query images. Combining two types of above approaches, \cite{sattler2015hyperpoints, sattler2012image} improve the structure-based methods by restricting features only matched to the 3D points that are visible in the top retrieved images. In this paper, we also incorporate these two types of methods to ensure both accuracy and efficiency.

When query and database images are taken under similar scenes and conditions, existing structure-based methods for visual localization tend to work well. However, since they rely heavily on local feature descriptors, which are vulnerable to the changes in illumination, weather, etc., it has been becoming a bottleneck restricting the robustness of existing localization algorithms. When images are taken under significantly different conditions or far apart in time, existing methods often fail to recover camera poses because feature descriptors are drastically changed. Fortunately, the high-level semantic information of image is more robust and invariant than the underlying local image features. Therefore, it is reasonable to fuse the underlying textural information and high-level semantic information for visual localization.

In this paper, we propose a robust method for visual localization by utilizing semantic information. As the semantic information is comparatively invariant under different conditions, it can be used as a supervisor to distinguish correct retrieved images from all retrieved database images. At first, we obtain the 3D model by running SfM algorithm on all database images and assign each 3D model point a semantic label. For a query image, we compute its semantic segmentation and search its top-$k$ similar database images. Then, we compute the query image's temporary pose using 2D-3D matches produced by matches between the query image and each one of the retrieved images, and assign semantic consistency scores for these 2D-3D matches belonging to the selected retrieved image. Finally, we put all 2D-3D matches together with their semantic consistency scores as weights and run a weighted RANSAC-based PnP solver to recover the final query pose. Thanks to the usage of high-level semantic information in both 3D model and the query image, the proposed method could achieve robust and accurate visual localization results on benchmark dataset \cite{sattler2018benchmarking}.

Compared to recent semantic localization approaches \cite{8578819, toft2017long, toft2018semantic}, our paper makes the following contributions: 1) We propose a new localization pipeline that incorporates structure-based method and image-based method while utilizing semantic information at the same time. 2) We do not need any additional restrictions (known camera height and gravity direction from ground truth) compared with the state-of-the-art semantic visual localization method \cite{toft2018semantic}.

\section{Related work}
\label{sec:RW}

\noindent\textbf{Localization with semantics.} Recently, several visual localization methods using semantic information have been proposed. Sch{\"o}nberger \textit{et al.} \cite{8578819} use generative model to learn descriptors and the model was trained to complete the semantic scene. Learned descriptors are used for establishing 3D-3D matches and estimating an alignment to define the query image pose. Toft \textit{et al.} \cite{toft2017long} use optimization method to refine pose estimation by improving semantic consistency of the curve segments and the projected 3D points. The most similar work to ours is \cite{toft2018semantic}. Toft \textit{et al.} measure semantic consistency for each 2D-3D correspondence by projecting 3D semantic points into the query semantic images. Yet, they use gravity direction and camera height as prior knowledge which is not required in our method.

\noindent\textbf{Localization benchmarks.} Although notable datasets such as North Campus Long-Term (NCLT) \cite{carlevaris2016university} and KITTI \cite{geiger2013vision} provide visual localization benchmark with images captured over long period, they do not contain large viewing condition changes or only few scenes are visited multiple times. Recently, Sattler \textit{et al.} \cite{sattler2018benchmarking} create a benchmarking    RobotCar Seasons dataset for long-term visual localization under all kinds of challenging conditions including day-night changes, illumination changes (dawn/noon), as well as weather (dust/sun/rain/snow) and seasonal (winter/summer) variations. They manually labelled 2D-3D matches in some tough cases to overcome the impact of large condition variations for obtaining the ground truth. Therefore, we evaluate our method on this dataset.

\section{localization Using semantic 3D map}
\label{sec:method}

The main bottleneck of feature based visual localization methods is that they are fragile under large condition variations in lighting, weather, season, etc. By contrast, semantic information is comparatively invariant under different conditions. In this paper, we propose a new localization pipeline and we do not need any strict prior as \cite{toft2018semantic} did. The pipeline of our method is illustrated in Fig.\ref{fig:1}: 1) We first run a standard SfM algorithm to construct a sparse 3D model of the scene. 2) Given semantic segmentation about each database image, every 3D point can be assigned a semantic label so that the 3D model becomes a sparse semantic 3D map  $ M_{S} $. 3) We then use image retrieval method to get a set of candidate database images $ \textbf{I}_{\textbf{R}}=\lbrace I_{R}^{i} , i=1,2,...,k  \rbrace$ for the query image $ I_{Q} $. 4) For each pair of $ I_{Q} $ and $ I_{R}^{i} $, We establish 2D-3D matches between $ I_{Q} $ and $ M_{S} $ through 2D-2D feature matches indirectly. Using these matches we can recover a temporary camera pose for $ I_{Q} $ (related to $ I_{R}^{i} $). 5) Given the estimated pose and semantic segmentation of $ I_{Q} $, all 3D model points are projected into $ I_{Q} $. We measure the semantic consistency between the 3D points and the projections on $ I_{Q} $, and use it as the weight for all the 2D-3D matches related to image $ I_{R}^{i} $. 6) Finally we use 2D-3D matches related to all retrieved images together with their consistency weights to bias sampling during RANSAC-based pose estimation.

\subsection{Sparse semantic 3D map}
\label{ssec:submethod1}
We run a regular SfM pipeline using all database images to construct a sparse 3D model of the scene. After SfM, the location as well as the visible images of each 3D point are obtained. Given semantic segmentation for all database images using off-the-shelf segmentation CNNs (DeepLabv3+ network \cite{deeplabv3plus2018} in this paper), we assign each 3D point a semantic label by maximum voting with reprojection pixel labels in all its visible images. By exploiting semantics, we can remove dynamic objects in the 3D point cloud such as person, car, bicycle, rider, bus and so on. Finally, we can obtain a cleaner sparse semantic 3D map $ M_{S} $ than the original one.

\begin{figure}[htb]
	\centering  
	\includegraphics[width=7cm,height=4cm]{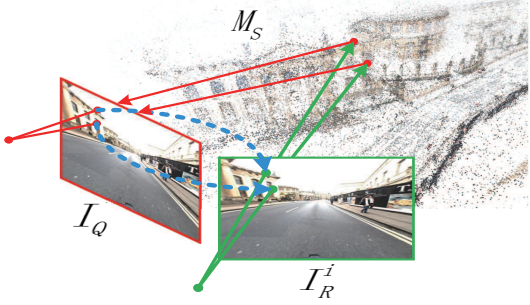}  
	\caption{Feature matching. The blue dotted lines are the feature matches between $I_{Q}$ and $ I_{R}^{i} $. The green solid lines are the 2D-3D correspondences between $ I_{R}^{i} $ and $ M_{S} $. The red solid lines are the 2D-3D matches between $ I_{Q} $ and $ M_{S} $. } 
	\label{fig:2}   
\end{figure}

\subsection{Semantic scores}
\label{ssec:submethod2}
Unlike \cite{toft2018semantic}, which measures the quality of 2D-3D matches directly through strict restrictions as prior, the main idea of our method is to measure the quality of each $ I_{R}^{i} $ without using any prior information.

We use NetVLAD \cite{arandjelovic2016netvlad} to obtain top-$k$ ranked database images $ \textbf{I}_{\textbf{R}} $ for each query image $I_{Q}$. The size of parameter $k$ depends on the computational efficiency and computing resources, and is set to 20-50 in our experiments. Due to the environment variance, erroneous retrieved images are inevitable. Fortunately, with the help of semantic information, we can alleviate the negative effect in most cases.

We use one $ I_{R}^{i} $ at a time for feature matching with $ I_{Q} $. Using KNN search and Lowe’s ratio test \cite{lowe2004distinctive}, 2D-2D matches (blue dotted lines in Fig.\ref{fig:2}) are computed. We use a relaxed ratio threshold of 0.9 to avoid rejecting possibly correct matches. The SfM model provides the 2D-3D correspondences (green solid lines in Fig.\ref{fig:2}) between the image points on $ I_{R}^{i} $ and the 3D points in $ M_{S} $. Therefore, we can obtain the 2D-3D matches (red solid lines in Fig.\ref{fig:2}) between $ I_{Q} $ and $ M_{S} $ through the 2D-2D matches and the SfM model. Then, these 2D-3D matches are used to recover a temporary query pose by applying a PnP solver. Given semantic segmentation about  $ I_{Q} $, we project all 3D points into $ I_{Q} $ by the estimated temporary query pose to check the number of consistent semantic points. Before we project all 3D points, we need to handle the situations of occlusion.

Apparently, not all 3D points are visible for a camera. Hence, we should only project the visible 3D points into the query image $ I_{Q} $. Similar to \cite{toft2018semantic}, we only consider the 3D point $\textbf{X}$ that satisfies:
\begin{equation}
\label{equ:1}　
d_{l}<\Vert \textbf{v}\Vert<d_{u} ,  \angle(\textbf{v},\textbf{v}_{m})<\theta   .
\end{equation}

\noindent where $ \textbf{v}= \textbf{C}_{Q}-\textbf{X}$. $\textbf{C}_{Q}$ is the estimated camera center of $ I_{Q} $.  $d_{l}$ denotes the minimum distance between the 3D point $ \textbf{X}$ and the position of all database cameras which could observe it, while $d_{u}$ denotes the maximum distance. $\theta$ is the angle between the two extreme viewpoints from which the 3D point $\textbf{X}$ was triangulated and the unit vector $\textbf{v}_{m}$ is in the middle between the two extreme viewpoints. This means that 3D points used for projection should be seen by the query image from similar distance and direction as the 3D points are viewed from database images in the SfM procedure.

We count the number of 3D points whose labels are the same as their projections in the query image. We use the number as the semantic score of $ I_{R}^{i} $. Since we use all 3D model points, not only the 2D-3D matches, the quality of the semantic scores can be ensured. A high semantic score means the pose estimated by $ I_{R}^{i} $ tends more likely to be correct from the perspective of projection semantic consistency. In other words, the retrieved database images with high semantic scores can be considered as correct retrieved images, while those with low scores means erroneous retrieved images to some extent. In this way, each $ I_{R}^{i} $ can obtain a semantic score during the above procedure independently.

\begin{table*}
\small
\caption{Evaluation on the long-term visual localization dataset (RobotCar Seasons)}
\centering
\resizebox{\textwidth}{!}{
\begin{tabular}{|c|c|c|c|c|c|c|c||c|c|}
\hline
\multirow{2}*{ } & \multicolumn{7}{c||}{\textbf{Day conditions}} & \multicolumn{2}{c|}{\textbf{Night conditions}}\\
\cline{2-10}
&dawn&dusk&OC-summer&OC-winter&rain&snow&sun&night&night-rain \\
\cline{2-10}
m & { .25/.50/5.0} & { .25/.50/5.0}& { .25/.50/5.0}& { .25/.50/5.0}& { .25/.50/5.0}& { .25/.50/5.0}& { .25/.50/5.0 }&  { .25/.50/5.0}& { .25/.50/5.0}\\
deg & { 2/5/10} & { 2/5/10}& { 2/5/10}& { 2/5/10}& { 2/5/10}& { 2/5/10}&{  2/5/10}& { 2/5/10}& { 2/5/10}\\
\hline

{ ActiveSearch} & { 36.2/68.9/89.4} & { 44.7/74.6/95.9} &{ 24.8/63.9/95.5} &{ 33.1/71.5/93.8} &{ 51.3/79.8/96.9} & { 36.6/72.2/93.7} & { 25.0/46.5/69.1} &{  0.5/1.1/3.4} & { 1.4/3.0/5.2}\\

\hline
{ CSL} & { 47.2/73.3/90.1} & { 56.6/82.7/\textbf{95.9}} &{ 34.1/71.1/93.5} &{ 39.5/75.9/92.3} &{ 59.6/83.1/97.6} & { 53.2/83.6/92.4} & { 28.0/47.0/70.4} &{  0.2/0.9/5.3} & { 0.9/4.3/9.1}\\

\hline

{ DenseVLAD} & { 8.7/36.9/92.5} & { 10.2/38.8/94.2} &{ 6.0/29.8/92.0} &{ 4.1/26.9/93.3} &{ 10.2/40.6/96.9} & { 8.6/30.1/90.2} & { 5.7/16.3/80.2} &{ 0.9/3.4/19.9} & { 1.1/5.5/25.5}\\

\hline

{ NetVLAD} & { 6.2/22.8/82.6} & { 7.4/29.7/92.9} &{ 6.5/29.6/95.2} &{ 2.8/26.2/92.6} &{ 9.0/35.9/96.0} & { 7.0/25.2/91.8} & { 5.7/16.5/86.7} &{ 0.2/1.8/15.5} & { 0.5/2.7/16.4}\\

\hline

{ FABMAP} & { 1.2/5.6/14.9} & { 4.1/18.3/55.1} &{ 0.9/8.9/39.3} &{ 2.6/13.3/44.1} &{ 8.8/32.1/86.5} & { 2.0/8.2/28.4} & { 0.0/0.0/2.4} &{ 0.0/0.0/0.0} & { 0.0/0.0/0.0}\\

\hline

{Non-semantics} & { 32.5/58.6/82.6} & { 38.3/71.8/94.2} &{ 23.5/56.2/85.5} &{ 27.9/64.1/87.2} &{ 49.9/77.2/96.9} & { 34.4/65.2/83.6} & { 13.2/25.2/45.0} &{ 0.0/0.0/1.8} & { 0.0/0.0/2.0}\\

\hline

{  Ours} & { \textbf{54.2}/\textbf{79.9}/\textbf{95.2}} & { \textbf{60.9}/\textbf{83.2}/95.7} &{ \textbf{43.8}/\textbf{79.5}/\textbf{99.1}} &{ \textbf{46.9}/\textbf{82.1}/\textbf{96.2}} &{ \textbf{62.9}/\textbf{84.1}/\textbf{97.6}} & { \textbf{61.1}/\textbf{85.9}/\textbf{96.1}} & { \textbf{50.7}/\textbf{76.3}/\textbf{95.4}} &{ \textbf{9.1}/\textbf{21.2}/\textbf{41.3}} & { \textbf{13.9}/\textbf{32.0}/\textbf{43.6}}\\

\hline

\end{tabular}
\label{tab1}}
\end{table*}

\begin{table}
\footnotesize
\caption{Comparison with \cite{toft2018semantic}}
\centering
\setlength{\tabcolsep}{7mm}{
\begin{tabular}{|c|c|}

\hline
 & all day \\
 
\cline{2-2}

m &  .25/.50/5.0  \\

deg &  2/5/10  \\

\hline
{ Semantic Match Consistency} & 50.6/79.8/95.1 \\
\hline
Ours & \textbf{54.4}/\textbf{81.5}/\textbf{96.5}  \\
\hline

\end{tabular}
\label{tab2}}
\end{table}

\subsection{Weighted RANSAC pose estimation}
\label{ssec:submethod3}
Finally, we put all 2D-3D matches produced by each $ I_{R}^{i} $ and $ I_{Q} $ together to run a final PnP solver, inside a RANSAC loop. The 2D-3D matches produced by the same $ I_{R}^{i} $ are assigned a same score which equals to the semantic score of $ I_{R}^{i} $. We normalize each score by the sum of scores of all 2D-3D matches and use the normalized score as a weight $ p $ for RANSAC's sampling. A 2D-3D match is selected with the probability $ p $ inside the RANSAC loop. This means if $ I_{R}^{i} $ has a high semantic score (correct retrieved image), the 2D-3D matches produced by $ I_{R}^{i} $ will be selected with the same high probability in the RANSAC loop . Compared with directly removing 2D-3D matches with low semantic scores, this semantic weighted RANSAC strategy guarantees that we only use semantic information as a soft constraint and makes our approach more robust in semantically ambiguous situations.

\section{Experimental Evaluation}
\label{sec:exper}

We evaluate the proposed method on benchmark visual localization dataset RobotCar Seasons \cite{sattler2018benchmarking}. In the following, we will introduce the dataset and make a detailed comparison with other existing approaches.

The long-term visual localization RobotCar Seasons dataset \cite{sattler2018benchmarking} is based on a subset of the Oxford RobotCar dataset \cite{maddern20171}. It contains 20,862 database images and 11,934 query images, which covers a wide range of environmental condition variations including season changes, weather variations, even day-night changes. Besides, some images contain motion blur which causes more difficulties for accurate visual localization.

The SfM model of the dataset is provided by \cite{sattler2018benchmarking} which is produced by state-of-the-art opensource 3D system COLMAP \cite{schonberger2016structure}. To augment this SfM model with semantics, we use DeepLabv3+ network \cite{deeplabv3plus2018} to segment all dataset images and assign a label to each 3D point by maximum voting with reprojection pixel labels in all its visible database images. The DeepLabv3+ network \cite{deeplabv3plus2018} is pre-trained on the Cityscapes \cite{cordts2016cityscapes} dataset. Additionally, we manually annotate 20 night condition images from the origin RobotCar dataset \cite{maddern20171} and use them to fine-tune the pre-trained model in order to improve the segmentation performance for all conditions. The semantic classes we use are the same as Cityscapes.

In the image retrieval step, we use NetVLAD \cite{arandjelovic2016netvlad} and the pre-trained Pitts30K model to generate 4096-dimensional descriptor vectors for each query and database image. Then, normalized L2 distances of the descriptors are computed, the top-$k$ best database matching images are chosen as candidate images. We set the retrieval number, $k$, to 30 in day conditions and 50 in night conditions. We record the percentage of query images which are localized within $X$m and $Y^{\circ}$  compared to ground truth. The same as \cite{sattler2018benchmarking}, we use three different thresholds: ($ 0.25 $m, $2^{\circ}$), ($0.5$m, $ 5^{\circ}$) and ($5$m, $10^{\circ}$), representing high, medium and coarse precision respectively.

To verify the effectiveness of the semantic information, we first make a comparative experiment. The \textit{Non-semantics} in Table \ref{tab1} is our approach without using semantics. It uses 2D-3D matches produced by all retrieved database images to run a PnP solver and have identical selection probability in the RANSAC loop. As we can see in Table \ref{tab1}, by exploiting semantics our method leads to significant improvement of localization performance than the method without utilizing semantics. Then, we compare our method against some state-of-the-art approaches, using the results from \cite{sattler2018benchmarking, toft2018semantic} directly. We compare with two VLAD-based image retrieval methods, namely DenseVLAD \cite{torii201524} and NetVLAD \cite{arandjelovic2016netvlad}. FAB-MAP \cite{cummins2008fab} is also a image retrieval approach based on the Bag-of-Words (BoW) paradigm \cite{sivic2003video}. As for structure-based algorithms, we compare with \textit{Active Search} \cite{sattler2017efficient} and \textit{City-Scale Localization} (CSL) \cite{svarm2017city}. As current learning-based methods can not achieve competitive performance \cite{sattler2018benchmarking}, we do not incorporate this type of methods for further comparison. As shown in Table \ref{tab1}, our method significantly outperforms all state-of-the-art approaches, which indicates that semantics could significantly improve the robustness of visual localization in our method. Our approach is only slightly inferior to CSL in coarse precision, and the reason for this is likely due to the results of image retrieval. In some cases, the environment where the query images were taken in has huge differences from that of the database images. In these situations, only few correct database images are retrieved in the top-$k$ retrieved images so that we can not locate camera accurately.

In addition, we compare our method with \cite{toft2018semantic} which also uses semantics for localization. \cite{toft2018semantic} uses camera height and gravity direction as prior. In \cite{toft2018semantic}, gravity direction is extracted from the ground truth pose and camera height is obtained from the intersection of the database trajectory and the cone of possible poses. In contrast, our method does not require any priors. As shown in Table \ref{tab2}, even without prior information our method still outperforms \cite{toft2018semantic} in all of conditions. 

\section{Conclusion}
\label{sec:majhead}

In this paper, we proposed a 6DoF visual localization method based on sparse 3D semantic model. By exploiting semantic information, we score each retrieved image by projecting all visible 3D points into query image and measuring the projection semantic consistency. The semantic consistency score is used as weight for all 2D-3D matches produced by the current retrieved image. We put 2D-3D matches related to all retrieved images together to run a final PnP solver inside a weighted RANSAC loop and obtain the final query image pose consequently. 

Experiments on the benchmark visual localization dataset shows that our method outperforms state-of-the-arts in the high and medium precision. Besides, compared with state-of-the-art semantic visual localization method \cite{toft2018semantic}, the proposed method does not require any prior information, which makes our method more practical. In the future, we would like to improve the image retrieval algorithm and attempt to replace the sparse model with dense 3D semantic model.

\footnotesize
\bibliographystyle{IEEEbib}
\bibliography{Template}

\end{document}